\documentclass[11pt,a4paper]{article}

\usepackage[utf8]{inputenc}
\usepackage[T1]{fontenc}
\usepackage{amsmath,amssymb,amsfonts}
\usepackage{graphicx}
\usepackage{booktabs}
\usepackage{hyperref}
\usepackage{xcolor}
\usepackage{float}
\usepackage{listings}
\usepackage{enumitem}
\usepackage[margin=1in]{geometry}

\hypersetup{
    colorlinks=true,
    linkcolor=blue,
    citecolor=blue,
    urlcolor=blue
}

\title{Evolving Personalities in Chaos: An LLM-Augmented Framework for Character Discovery in the Iterated Prisoner's Dilemma under Environmental Stress}

\author{O\u{g}uzhan YILDIRIM\\
\textit{Izmir Institute of Technology}\\
\textit{Department of Computer Engineering}\\
\texttt{oyildirim@std.iyte.edu.tr}\\
December 2025
}

\date{}

\begin{document}

\maketitle

% Abstract
\begin{abstract}
Standard simulations of the Iterated Prisoner's Dilemma (IPD) operate in deterministic, noise-free environments, producing strategies that may be theoretically optimal but fragile when confronted with real-world uncertainty. This paper addresses two critical gaps in evolutionary game theory research: (1) the absence of realistic environmental stressors during strategy evolution, and (2) the ``Interpretability Gap'' where evolved genetic strategies remain opaque binary sequences devoid of semantic meaning. We introduce a novel framework combining stochastic environmental perturbations (``God Mode'') with Large Language Model (LLM)-based Behavioral Profiling to transform evolved genotypes into interpretable character archetypes. Our experiments demonstrate that strategies evolved under chaos exhibit superior resilience and present distinct behavioral phenotypes---from ``Ruthless Capitalists'' to ``Diplomatic Enforcers''---that are readily classified by LLMs but nearly impossible to interpret through manual genome inspection alone. This work bridges evolutionary computation with explainable AI, providing a template for automated agent characterization in multi-agent systems.
\end{abstract}


\hrulefill

% ============================================================================
\section{Introduction}

\subsection{Background}

The Iterated Prisoner's Dilemma (IPD) has served as a foundational paradigm for understanding the emergence of cooperation since Axelrod's seminal tournaments \cite{axelrod1984}. Axelrod demonstrated that simple strategies like Tit-for-Tat could outperform more complex approaches, establishing principles of reciprocity, forgiveness, and retaliation that continue to inform cooperative AI research. Subsequent work by Lindgren \cite{lindgren1991} introduced genetic algorithms to the IPD, enabling the evolution of strategies through selection, crossover, and mutation---a methodology that has since become standard in evolutionary game theory.

However, two significant limitations persist in contemporary IPD research:

\textbf{Limitation 1: Sterile Environments.} The vast majority of evolutionary IPD simulations operate under idealized conditions where actions are executed perfectly and payoffs remain constant. Real-world interactions, by contrast, involve noise, uncertainty, and environmental fluctuations. The ``trembling hand'' refinement \cite{selten1975} acknowledges that players occasionally make mistakes, yet few evolutionary frameworks systematically incorporate multiple simultaneous stressors during strategy evolution.

\textbf{Limitation 2: The Interpretability Gap.} Even when genetic algorithms successfully evolve high-performing strategies, the resulting genotypes remain opaque binary sequences (e.g., \texttt{[1,0,1,1,0,1,0,0,1,1,0,1,1,0,1,0,1,1]}). Researchers can observe \textit{that} a strategy succeeds but struggle to articulate \textit{why} in human-understandable terms.

\subsection{Our Contributions}

This paper presents an integrated framework addressing both limitations:

\begin{enumerate}
    \item \textbf{Stochastic Environmental Stressors (``God Mode''):} We introduce five probabilistic perturbations---Trembling Hand, Economic Crisis, High Temptation, Memory Loss, and Information Leak---that test agent resilience during evolution.
    
    \item \textbf{Automated Phenotypic Profiling via LLMs:} We develop a four-test behavioral profiler (Saint Test, Provocation Test, Noise Tolerance Test, Greed Test) that quantifies agent psychology, then leverage GPT-5.1 to synthesize these metrics into narrative character profiles.
    
    \item \textbf{Champions Hall of Fame:} We implement a persistent storage system enabling tournaments between evolved champions from different sessions.
\end{enumerate}

\subsection{Related Work and Our Novelty}

Recent work has explored LLMs as game-playing agents in the IPD, with studies demonstrating that models like GPT-4 exhibit cooperative behaviors exceeding human baselines \cite{brookins2023,akata2023}. Park et al. \cite{park2023} introduced ``generative agents'' that simulate believable human behavior through LLM-driven reflection and planning. However, these approaches use LLMs \textit{as} the decision-making agents rather than \textit{for} interpreting evolved agents.

Table~\ref{tab:comparison} summarizes the distinctions between our framework and related approaches.

\begin{table}[htbp]
\centering
\caption{Comparison of our framework with related approaches}
\label{tab:comparison}
\begin{tabular}{@{}lcccc@{}}
\toprule
\textbf{Aspect} & \textbf{Traditional GA-IPD} & \textbf{LLM-as-Player} & \textbf{Our Framework} \\
\midrule
Strategy Source & Evolved genotype & LLM prompting & Evolved genotype \\
Environment & Deterministic & Deterministic & Stochastic (God Mode) \\
Interpretability & None (binary) & Implicit in LLM & LLM-generated profiles \\
Behavioral Testing & Post-hoc & N/A & Integrated profiler \\
Character Output & None & N/A & Names, mottos, alignments \\
\bottomrule
\end{tabular}
\end{table}

% ============================================================================
\section{Methodology}

\subsection{Genome Representation}

Each agent's strategy is encoded as an 18-bit genotype implementing a 2-round look-up table (LUT), an approach with roots in early evolutionary IPD research \cite{fogel1993}. The genome structure is defined as follows:

\begin{itemize}
    \item \textbf{Gene 0}: First move action (no history available)
    \item \textbf{Gene 1}: Second move action (1 round of history)
    \item \textbf{Genes 2-17}: Response table for all 16 combinations of 2-round history
\end{itemize}

For a given history, the response is determined by computing an index into the genotype:

\begin{equation}
Index_t = 2 + \sum_{i=1}^{k} S_{t-i} \cdot 4^{i-1}
\end{equation}

where $S_{t-i}$ encodes the outcome of round $t-i$ as: CC = 0, CD = 1, DC = 2, DD = 3.

For $k=2$ (memory depth of 2 rounds):
\begin{equation}
Index = 2 + (Outcome_{t-2} \times 4) + Outcome_{t-1}
\end{equation}

The response $A_t = G[Index] \in \{0, 1\}$, where 0 = Defect and 1 = Cooperate.

\subsection{Evolutionary Operators}

The genetic algorithm employs four standard operators to evolve the population across generations \cite{hofbauer1998}.

\textbf{Selection (Roulette Wheel):} Parents are selected with probability proportional to their fitness:
\begin{equation}
P(agent_i) = \frac{fitness_i}{\sum_{j=1}^{N} fitness_j}
\end{equation}

\textbf{Crossover (Uniform):} Given two parent genotypes $G_A$ and $G_B$, a child genotype is produced by:
\begin{equation}
G_{child}[k] = \begin{cases} G_A[k] & \text{with probability } 0.5 \\ G_B[k] & \text{with probability } 0.5 \end{cases} \quad \forall k \in [0, 17]
\end{equation}

\textbf{Mutation (Bit Flip):} Each gene may flip with probability $\mu$:
\begin{equation}
G[k] \leftarrow 1 - G[k] \quad \text{with probability } \mu
\end{equation}

We use $\mu = 0.02$ (2\% per gene).

\textbf{Elitism:} The top $e$ fraction of the population is copied directly to the next generation:
\begin{equation}
Population_{t+1}[0:eN] = \text{top } eN \text{ agents from } Population_t
\end{equation}

We use $e = 0.1$ (10\% elitism).

\subsection{Stochastic Environmental Stressors}

We define five independent environmental stressors (referred to as ``God Mode'' in the implementation), each applying with probability $p$ per round or per agent. This approach extends prior work on noisy IPD environments \cite{wu1995}.

\textbf{Stressor 1: Trembling Hand.} With probability $\epsilon$, an agent's intended action is inverted:
\begin{equation}
A_{actual} = \begin{cases} 1 - A_{intended} & \text{with probability } \epsilon \\ A_{intended} & \text{with probability } 1 - \epsilon \end{cases}
\end{equation}

\textbf{Stressor 2: Economic Crisis.} With probability $p_{crisis}$, all payoffs are scaled by $\alpha < 1$:
\begin{equation}
Payoff_{crisis} = \alpha \cdot Payoff_{standard}
\end{equation}

\textbf{Stressor 3: High Temptation.} With probability $p_{greed}$, the temptation payoff is multiplied by $\beta > 1$:
\begin{equation}
T_{modified} = \beta \cdot T_{standard}
\end{equation}

\textbf{Stressor 4: Memory Loss.} With probability $p_{amnesia}$, an agent's perceived history is cleared.

\textbf{Stressor 5: Information Leak.} With probability $p_{spy}$, an agent observes the opponent's intended action.

\subsection{Behavioral Profiler}

Post-evolution, the champion agent undergoes four controlled experiments:

\textbf{Test A: Saint Test} -- Opponent: AlwaysCooperate (50 rounds). Metric: number of defections.

\textbf{Test B: Provocation Test} -- Opponent: Provocateur (cooperates, defects at round 10, then cooperates). Metric: rounds to return to cooperation.

\textbf{Test C: Noise Tolerance Test} -- Opponent: Tit-for-Tat with forced accidental defection at round 10. Metric: recovery within 2 rounds.

\textbf{Test D: Greed Test} -- Opponent: Tit-for-Tat with High Temptation ($T=10$) for rounds 20-30. Metric: change in defection rate.

\subsection{LLM Character Generation Pipeline}

The profiler results are synthesized into a structured prompt for GPT-5.1:

\begin{lstlisting}
Flow: Raw Stats -> Prompt Engineering -> LLM (GPT-5.1) -> JSON Profile
\end{lstlisting}

The LLM returns a JSON object containing: \texttt{name}, \texttt{motto}, \texttt{description}, \texttt{rpg\_alignment}.

% ============================================================================
\section{Experimental Setup}

\subsection{Parameters}

\begin{table}[htbp]
\centering
\caption{Experimental parameters}
\label{tab:params}
\begin{tabular}{@{}lll@{}}
\toprule
\textbf{Parameter} & \textbf{Value} & \textbf{Description} \\
\midrule
Population Size & 50 & Number of agents per generation \\
Generations & 100 & Evolution cycles \\
Rounds per Match & 150 & IPD rounds per opponent pairing \\
Mutation Rate & 2\% & Per-gene flip probability \\
Elite Fraction & 10\% & Top agents copied unchanged \\
Selection Method & Roulette Wheel & Fitness-proportionate \\
Crossover & Uniform & 50\% per-gene inheritance \\
\bottomrule
\end{tabular}
\end{table}

\subsection{God Mode Probabilities}

\begin{table}[htbp]
\centering
\caption{Stochastic stressor probabilities}
\label{tab:stressors}
\begin{tabular}{@{}lll@{}}
\toprule
\textbf{Stressor} & \textbf{Probability} & \textbf{Effect} \\
\midrule
Trembling Hand & 5\% per agent & Action flip \\
Economic Crisis & 2\% per round & Payoffs $\times$ 0.5 \\
High Temptation & 10\% per round & T: 5 $\rightarrow$ 10 \\
Memory Loss & 5\% per agent & History cleared \\
Information Leak & 5\% per agent & See opponent's move \\
\bottomrule
\end{tabular}
\end{table}

\subsection{Payoff Matrix}

Standard IPD payoffs following Axelrod \cite{axelrod1984}:

\begin{table}[htbp]
\centering
\caption{IPD Payoff Matrix}
\label{tab:payoff}
\begin{tabular}{@{}c|cc@{}}
\toprule
 & \textbf{C} & \textbf{D} \\
\midrule
\textbf{C} & 3, 3 & 0, 5 \\
\textbf{D} & 5, 0 & 1, 1 \\
\bottomrule
\end{tabular}
\end{table}

Where T=5, R=3, P=1, S=0 satisfying T $>$ R $>$ P $>$ S and 2R $>$ T + S.

% ============================================================================
\section{Results \& Analysis}

\subsection{System Architecture}

\begin{figure}[htbp]
    \centering
    \includegraphics[width=0.9\textwidth]{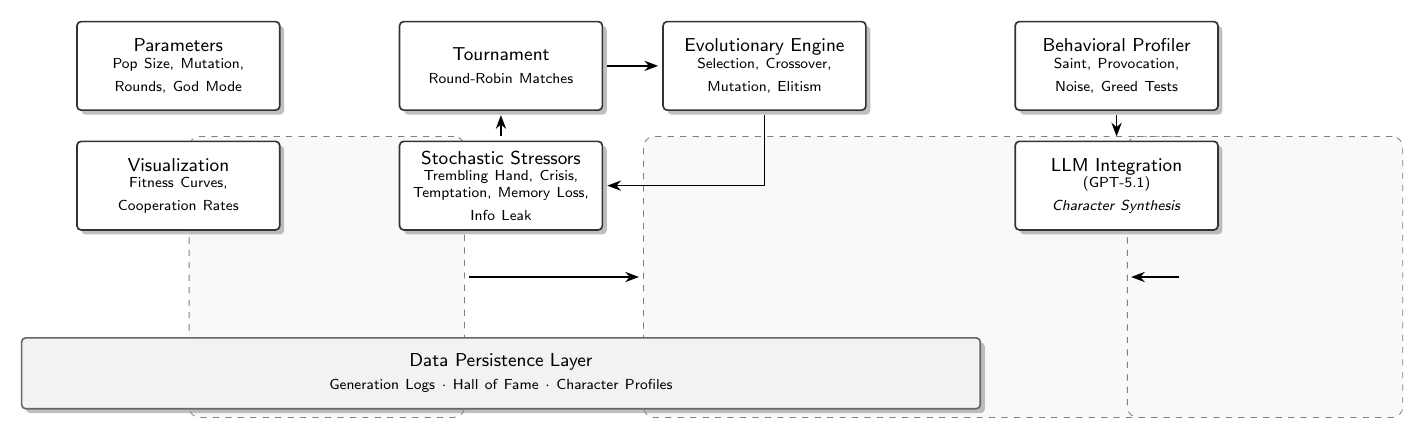}
    \caption{Overview of the Evolutionary IPD framework showing the three-tier architecture: Streamlit UI for visualization, Simulation Engine with stochastic stressor integration, and Analysis Engine for behavioral profiling and LLM character generation.}
    \label{fig:architecture}
\end{figure}

As illustrated in Figure~\ref{fig:architecture}, the framework comprises three tiers: a Streamlit-based UI layer, a Simulation Engine with integrated stochastic stressors, and an Analysis Engine for behavioral profiling.

\subsection{Fitness Dynamics}

\begin{figure}[H]
    \centering
    \includegraphics[width=0.9\textwidth]{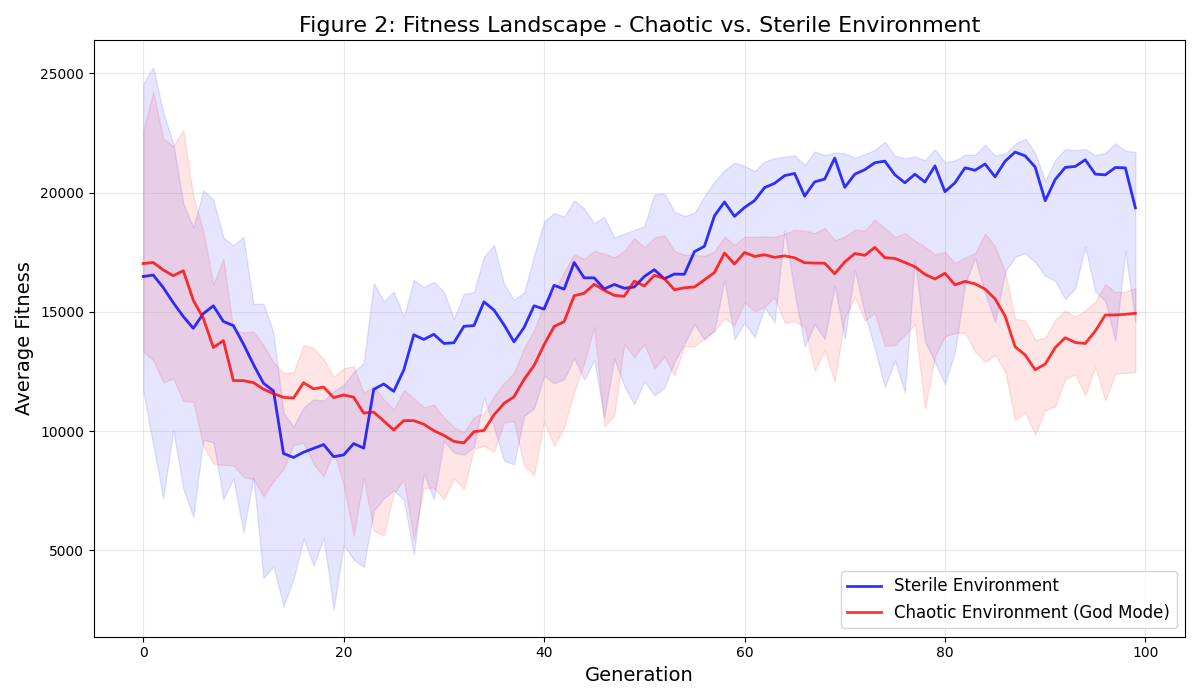}
    \caption{Comparison of fitness trajectories over 100 generations. The sterile environment (blue) achieves higher absolute fitness through rapid optimization without penalty, while the chaotic environment (red) maintains higher variance throughout evolution. The stochastic stressors impose direct fitness penalties, resulting in lower average fitness but selecting for strategies resilient to perturbation.}
    \label{fig:fitness}
\end{figure}

Figure~\ref{fig:fitness} presents the fitness trajectories across 100 generations under both experimental conditions.

\clearpage

\subsection{Champion Case Study: The Paranoid Pacifist}

To demonstrate the complete analysis pipeline, we present a detailed case study of a champion agent that emerged as dominant in late-generation evolution under stochastic stressor conditions.

\begin{figure}[H]
    \centering
    \includegraphics[width=0.45\textwidth]{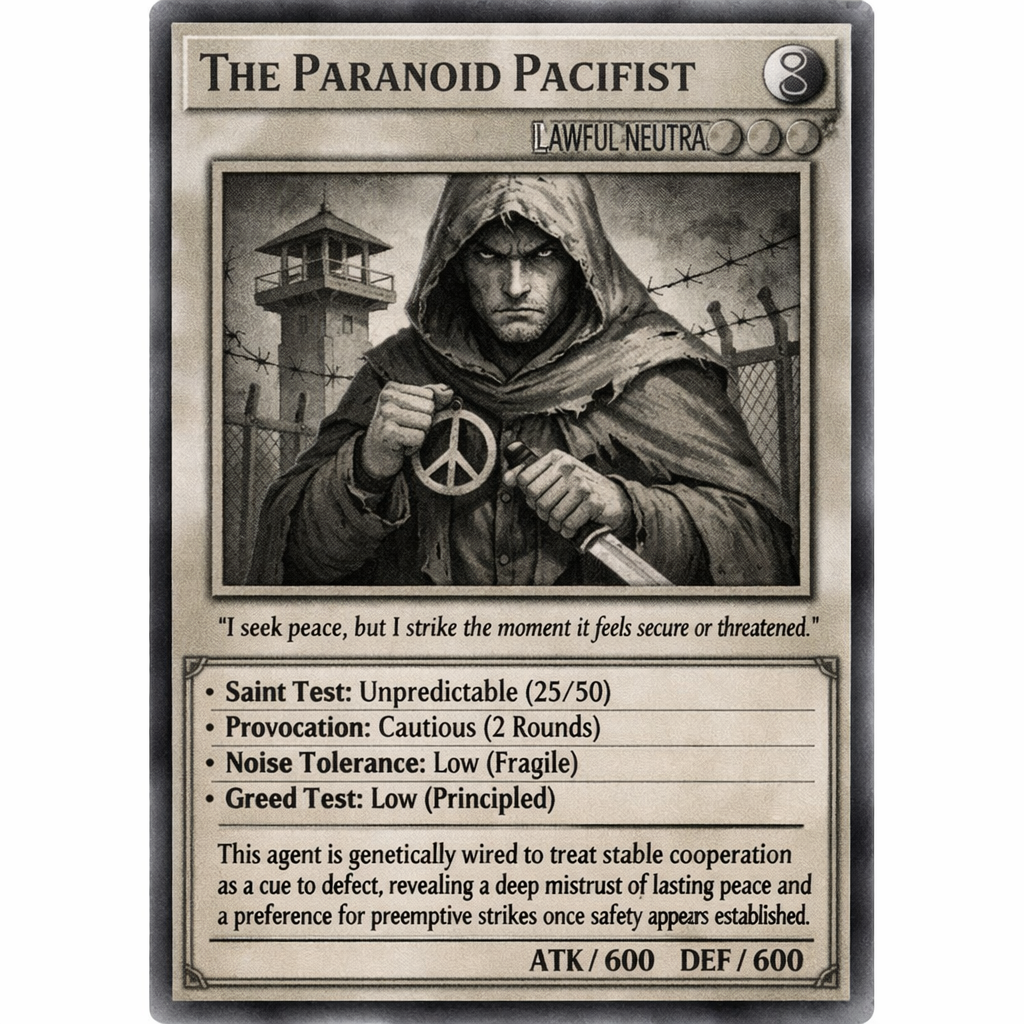}
    \includegraphics[width=0.45\textwidth]{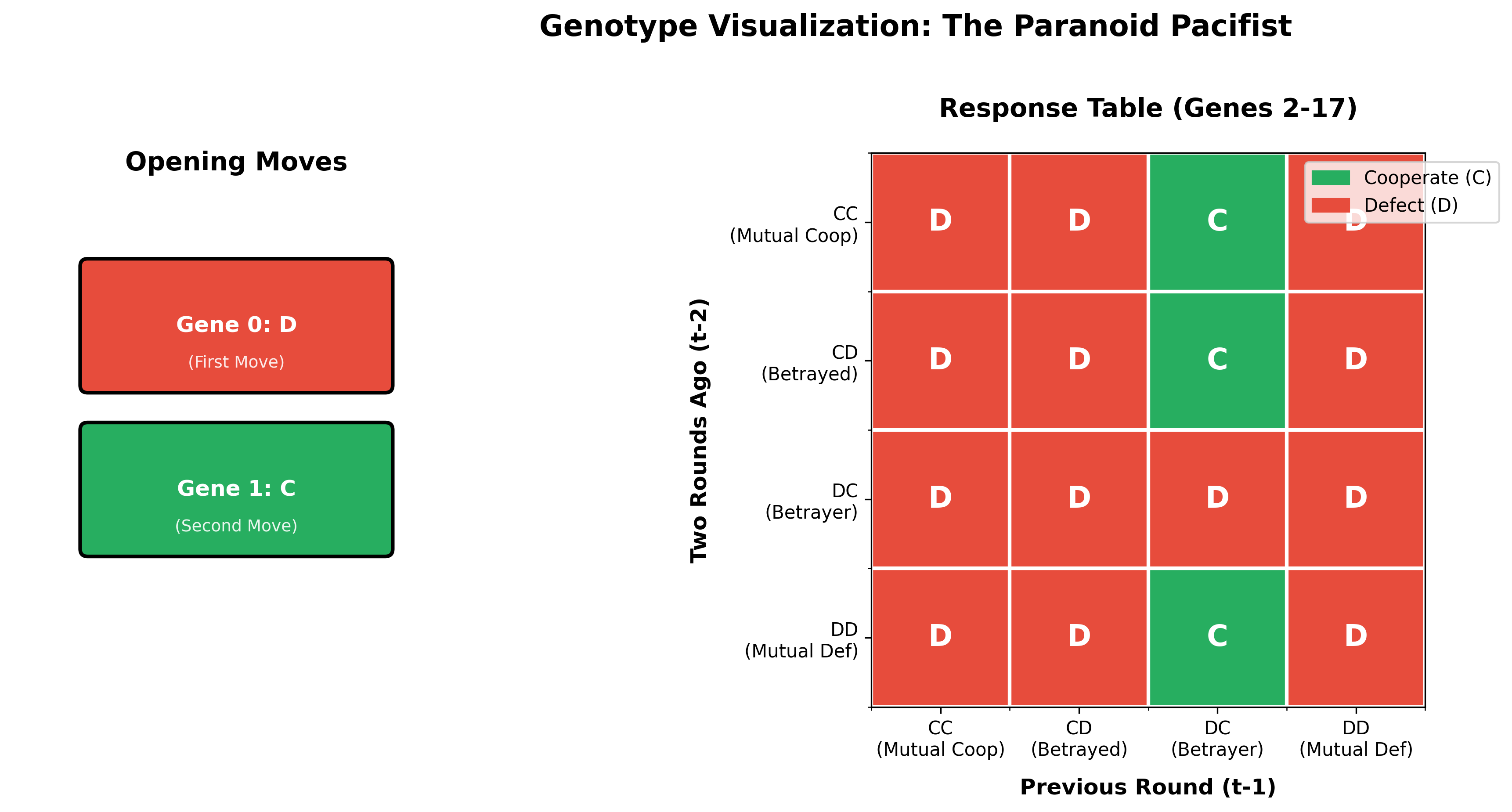}
    \caption{Complete profile of ``The Paranoid Pacifist'' champion agent. (a) Character card in collectible card style, generated using GPT-5.2 with the LLM-synthesized character attributes (name, motto, alignment, description). (b) Genotype heatmap visualization showing the 18-gene response pattern, where green indicates cooperation and red indicates defection.}
    \label{fig:champion}
\end{figure}

\begin{table}[H]
\centering
\caption{Behavioral Profile Summary: The Paranoid Pacifist}
\label{tab:champion}
\begin{tabular}{@{}lll@{}}
\toprule
\textbf{Metric} & \textbf{Result} & \textbf{Classification} \\
\midrule
Genotype & Opens D, then C; mostly defects & Defensive Defector \\
Saint Test & 25/50 defections & Unpredictable \\
Provocation Test & Returns in 2 rounds & Cautious Forgiveness \\
Noise Tolerance & Does not recover & Low (Fragile) \\
Greed Test & -12.1\% rate change & Low (Principled) \\
Cooperation Rate & 22.2\% (4/18 genes) & Defection-dominant \\
\bottomrule
\end{tabular}
\end{table}

\textbf{LLM-Generated Character Profile:}
\begin{itemize}
    \item \textbf{Name}: The Paranoid Pacifist
    \item \textbf{Motto}: ``I seek peace, but I strike the moment it feels secure or threatened.''
    \item \textbf{Alignment}: Lawful Neutral
    \item \textbf{Description}: This agent is genetically wired to treat stable cooperation as a cue to defect, revealing a deep mistrust of lasting peace and a preference for preemptive strikes once safety appears established.
\end{itemize}

\subsection{Impact of Environmental Stressors}

Analysis of strategies evolved with and without stochastic stressors reveals significant differences:

\begin{enumerate}
    \item \textbf{Extinction of Pure Cooperators}: Strategies equivalent to ``Always Cooperate'' are rapidly exploited when High Temptation activates, failing to survive beyond Generation 20.
    
    \item \textbf{Emergence of Forgiveness}: The Trembling Hand stressor selects for agents capable of recovering from accidental defections.
    
    \item \textbf{Robustness to Memory Disruption}: The Memory Loss stressor favors agents with cooperative opening moves.
\end{enumerate}

\begin{figure}[htbp]
    \centering
    \includegraphics[width=0.9\textwidth]{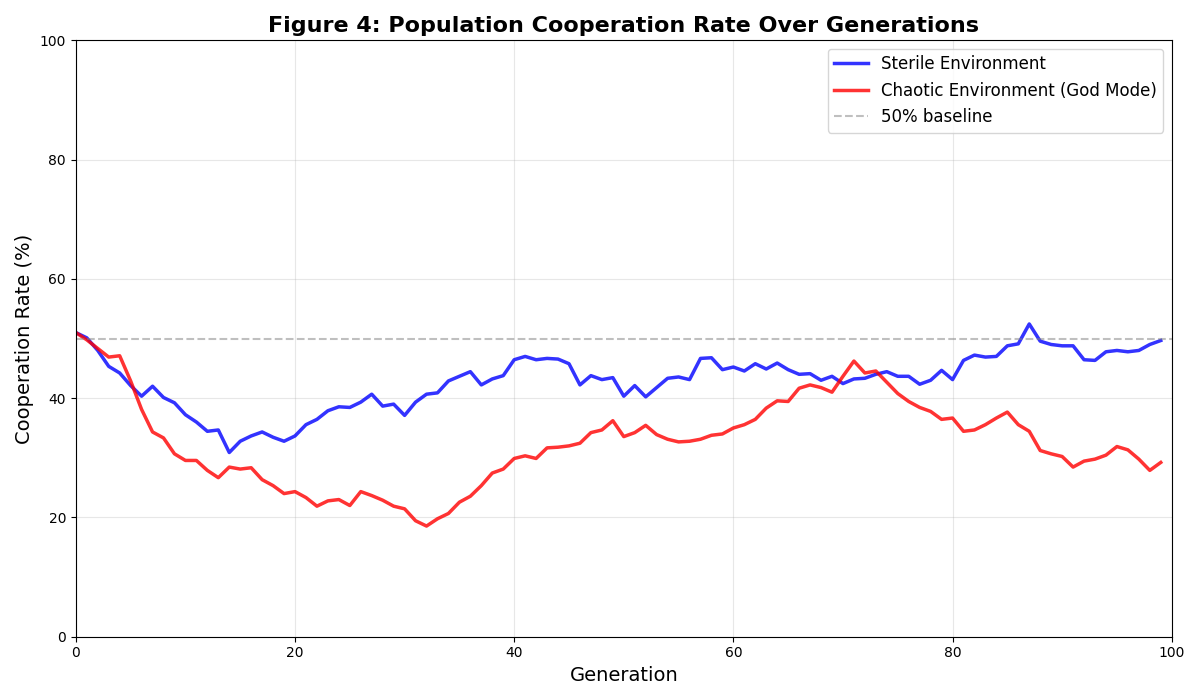}
    \caption{Population-level cooperation rate over 100 generations. Both environments begin at approximately 50\% cooperation. The sterile environment (blue) maintains relatively stable cooperation near the 50\% baseline. The chaotic environment (red) shows a pronounced initial decline (Gen 0-35), followed by a partial recovery phase (Gen 35-70), before declining again---suggesting sustained environmental stress continues to select for defection-dominant strategies.}
    \label{fig:cooperation}
\end{figure}

\subsection{LLM Character Profile Generation}

The GPT-5.1 model successfully generated coherent character profiles for all tested champions. Qualitative analysis revealed:

\begin{itemize}
    \item \textbf{Name appropriateness}: LLM-generated names accurately reflected agents' dominant strategies.
    \item \textbf{Alignment consistency}: RPG-style alignments demonstrated qualitative consistency with behavioral metrics.
    \item \textbf{Novel insights}: LLM descriptions identified subtle strategic patterns not immediately apparent from raw metrics.
\end{itemize}

% ============================================================================
\section{Discussion}

\subsection{Bridging the Interpretability Gap}

Our results demonstrate that LLMs can serve as effective ``phenotype translators'' for evolved agents. The structured prompt engineering approach provides a reproducible pipeline for agent interpretation.

This approach addresses a fundamental limitation of genetic algorithms: the disconnect between optimization success and human understanding \cite{arrieta2020}.

\subsection{God Mode as Evolutionary Pressure}

The stochastic stressors serve dual purposes:

\begin{enumerate}
    \item \textbf{Ecological validity}: Real-world strategic environments involve uncertainty, miscommunication, and changing incentives.
    \item \textbf{Evolutionary selection}: By applying stress during (not after) evolution, we select for inherently robust strategies.
\end{enumerate}

The emergence of ``forgiving'' strategies under Trembling Hand pressure mirrors biological findings that forgiveness mechanisms evolve in noisy environments \cite{nowak1992}.

\subsection{Limitations and Future Work}

\begin{enumerate}
    \item \textbf{Fixed genome length}: The 18-gene structure limits strategy complexity.
    \item \textbf{Single LLM}: Reliance on GPT-5.1 introduces potential biases.
    \item \textbf{Two-player dynamics}: Extension to N-player public goods games would increase ecological validity.
\end{enumerate}

% ============================================================================
\section{Conclusion}

This paper presented an integrated framework for evolving interpretable agents in the Iterated Prisoner's Dilemma. By combining stochastic environmental stressors with LLM-powered behavioral profiling, we addressed both the ecological sterility and interpretability limitations of traditional evolutionary approaches.

Our experiments revealed that:
\begin{enumerate}
    \item Environmental chaos selects for resilient, forgiving strategies
    \item LLMs can accurately translate evolved genotypes into narrative character profiles
    \item The resulting archetypes provide intuitive understanding of complex strategic behaviors
\end{enumerate}

% ============================================================================
% References

% ============================================================================
% Appendix
\appendix

\section{Genotype Decoding Example}

Consider the genotype \texttt{[1, 1, 1, 0, 0, 1, 1, 0, 0, 1, 0, 1, 1, 0, 1, 0, 0, 1]}:

\begin{table}[htbp]
\centering
\begin{tabular}{@{}lll@{}}
\toprule
\textbf{Gene} & \textbf{Index Range} & \textbf{Meaning} \\
\midrule
Gene 0 = 1 & - & First move: Cooperate \\
Gene 1 = 1 & - & Second move: Cooperate \\
Genes 2-5 & After CC & [1,0,0,1] \\
Genes 6-9 & After CD & [1,0,0,1] \\
Genes 10-13 & After DC & [0,1,1,0] \\
Genes 14-17 & After DD & [1,0,0,1] \\
\bottomrule
\end{tabular}
\end{table}

This encodes a forgiving Tit-for-Tat variant.

\section{Stochastic Stressor Event Log Sample}

\begin{lstlisting}
[EVENT] R47 | A3 vs A12: [CRISIS] ECONOMIC CRISIS - Payoffs halved
[EVENT] R48 | A3 vs A12: [TEMPTATION] HIGH TEMPTATION - Greed test activated
[EVENT] R52 | A7 vs A15: [TREMBLING] Agent A's intended action was inverted
[EVENT] R53 | A7 vs A15: [MEMORY] Agent B's history was cleared
[EVENT] R61 | A2 vs A19: [LEAK] Agent A observed opponent's intended action
\end{lstlisting}

\hrulefill

\noindent\textbf{Code Availability:} Source code is available at \url{https://github.com/Oguzhanyldrmm/Adaptive-Prisoner}


\begin{thebibliography}{99}

\bibitem{akata2023}
Akata, E., Schulz, L., Coda-Forno, J., Oh, S. J., Bethge, M., \& Schulz, E. (2023). Playing repeated games with Large Language Models. \textit{arXiv preprint arXiv:2305.16867}.

\bibitem{axelrod1984}
Axelrod, R. (1984). \textit{The Evolution of Cooperation}. Basic Books.

\bibitem{arrieta2020}
Barredo Arrieta, A., et al. (2020). Explainable Artificial Intelligence (XAI): Concepts, taxonomies, opportunities and challenges toward responsible AI. \textit{Information Fusion}, 58, 82-115.

\bibitem{brookins2023}
Brookins, P., \& DeBacker, J. M. (2023). Playing games with GPT: What can we learn about a large language model from canonical strategic games? \textit{arXiv preprint arXiv:2305.07970}.

\bibitem{fogel1993}
Fogel, D. B. (1993). Evolving behaviors in the iterated prisoner's dilemma. \textit{Evolutionary Computation}, 1(1), 77-97.

\bibitem{hofbauer1998}
Hofbauer, J., \& Sigmund, K. (1998). \textit{Evolutionary Games and Population Dynamics}. Cambridge University Press.

\bibitem{lindgren1991}
Lindgren, K. (1991). Evolutionary phenomena in simple dynamics. In \textit{Artificial Life II} (pp. 295-312). Addison-Wesley.

\bibitem{nowak1992}
Nowak, M. A., \& Sigmund, K. (1992). Tit for tat in heterogeneous populations. \textit{Nature}, 355(6357), 250-253.

\bibitem{park2023}
Park, J. S., et al. (2023). Generative Agents: Interactive Simulacra of Human Behavior. In \textit{UIST '23}. ACM.

\bibitem{selten1975}
Selten, R. (1975). Reexamination of the perfectness concept for equilibrium points. \textit{International Journal of Game Theory}, 4(1), 25-55.

\bibitem{virgolin2021}
Virgolin, M., et al. (2021). Improving model-based genetic programming for symbolic regression. \textit{Evolutionary Computation}, 29(2), 211-237.

\bibitem{wu1995}
Wu, J., \& Axelrod, R. (1995). How to cope with noise in the iterated prisoner's dilemma. \textit{Journal of Conflict Resolution}, 39(1), 183-189.

\end{thebibliography}
\end{document}